\documentclass[conference]{IEEEtran}

\usepackage{graphicx}
\usepackage{float}
\usepackage{enumitem}
\usepackage{makecell}
\usepackage{multirow}
\usepackage{arydshln} 

\usepackage{bm}
\usepackage{tikz}
\usepackage{comment}
\usepackage{amsmath,amssymb} % define this before the line numbering.

\usepackage[utf8]{inputenc} % allow utf-8 input
\usepackage[T1]{fontenc}    % use 8-bit T1 fonts
\usepackage{hyperref}       % hyperlinks
\usepackage{url}            % simple URL typesetting
\usepackage{booktabs}       % professional-quality tables
\usepackage{amsfonts}       % blackboard math symbols
\usepackage{nicefrac}       % compact symbols for 1/2, etc.
\usepackage{microtype}      % microtypography
\usepackage{xcolor}         % colors

\IEEEoverridecommandlockouts
\usepackage{cite}
\usepackage{amsmath,amssymb,amsfonts}
\usepackage{algorithmic}
\usepackage{graphicx}
\usepackage{textcomp}
\usepackage{xcolor}
\def\BibTeX{{\rm B\kern-.05em{\sc i\kern-.025em b}\kern-.08em
    T\kern-.1667em\lower.7ex\hbox{E}\kern-.125emX}}
\begin{document}

\title{DocMAE: Document Image Rectification via Self-supervised Representation Learning}
% *\\

% \thanks{This work was supported by the National Natural Science Foundation of China under Contract 62021001.}

\author{\IEEEauthorblockN{Shaokai Liu\IEEEauthorrefmark{2}}
\IEEEauthorblockA{\textit{University of Science and Technology} \\
\textit{of China}\\
Hefei, China \\
liushaokai@mail.ustc.edu.cn}
\and
\IEEEauthorblockN{Hao Feng\IEEEauthorrefmark{2}}
\IEEEauthorblockA{\textit{University of Science and Technology} \\
\textit{of China}\\
Hefei, China \\
haof@mail.ustc.edu.cn}
\IEEEcompsocitemizethanks{\IEEEcompsocthanksitem
\IEEEauthorrefmark{2} The first two authors contribute equal to this work.}
\and
\IEEEauthorblockN{Wengang Zhou}
\IEEEauthorblockA{\textit{University of Science and Technology} \\
\textit{of China}\\
% \textit{Hefei Comprehensive National Science Center}\\
Hefei, China \\
zhwg@ustc.edu.cn}
\and
\IEEEauthorblockN{Houqiang Li}
\IEEEauthorblockA{\textit{University of Science and Technology of China} \\
% \textit{of China}\\
Hefei, China \\
lihq@ustc.edu.cn}
\and
\IEEEauthorblockN{Cong Liu}
\IEEEauthorblockA{\textit{iFLYTEK Research} \\
% \textit{Research}\\
Hefei, China \\
congliu2@iflytek.com}
\and
\IEEEauthorblockN{Feng Wu}
\IEEEauthorblockA{\textit{University of Science and Technology of China} \\
% \textit{of China}\\
Hefei, China \\
fengwu@ustc.edu.cn}
\thanks{This work is supported by the Strategic Priority Research Program of Chinese Academy of Sciences under Grant No. XDC08030000.}
}

% \author{\IEEEauthorblockN{Shaokai Liu\IEEEauthorrefmark{1}}
% \textit{University of Science and Technology of China}\\
% \and
% \IEEEauthorblockN{Hao Feng\IEEEauthorrefmark{1}}
% \textit{University of Science and Technology of China}\\
% \and
% \IEEEauthorblockN{Wengang Zhou}
% \textit{University of Science and Technology of China}\\
% \and
% \IEEEauthorblockN{Houqiang Li}
% \textit{University of Science and Technology of China}\\
% \and
% \IEEEauthorblockN{Cong Liu}
% \textit{University of Science and Technology of China}\\
% congliu2@iflytek.com}
% \and
% \IEEEauthorblockN{Feng Wu}
% \textit{University of Science and Technology of China}\\
% fengwu@ustc.edu.cn}
% }

\maketitle

\begin{abstract}
Tremendous efforts have been made on document image rectification, but how to learn effective representation of such distorted images is still under-explored.
In this paper, we present DocMAE, a novel self-supervised framework for document image rectification. 
Our motivation is to encode the structural cues in document images by leveraging masked autoencoder to benefit the rectification,
$i.e.$, the document boundaries, and text lines. 
Specifically, we first mask random patches of the background-excluded document images and then reconstruct the missing pixels. 
With such a self-supervised learning approach, the network is encouraged to learn the intrinsic structure of deformed documents by restoring document boundaries and missing text lines.
Transfer performance in the downstream rectification task validates the effectiveness of our method.
Extensive experiments are conducted to demonstrate the effectiveness of our method.
% Our codes will be made publicly available.

\end{abstract}
%

% 待确认
% \begin{keywords}
% Document Image Rectification, Self-supervised Learning, Transformer
% \end{keywords}
%


\section{Introduction}
With the ubiquitous accessibility of smartphones, document digitization becomes much more convenient than before. 
However, document images captured by smartphones usually suffer from various distortions, due to some stochastic factors, such as sheet deformations, camera angles, and scene illuminations.
They bring difficulties to the downstream visual tasks, such as automatic text recognition~\cite{pozo2011method}, content analysis~\cite{wu2021document}, and question answering~\cite{yang2022modality}.
To overcome these issues, document image rectification has been actively studied in the past decades.

Traditional solutions~\cite{937649,4407722,6909892,4916075,you2017multiview,das2017common} to document image rectification are usually based on 3D reconstruction techniques. 
These methods either resort to extra hardware or register multi-view images to reconstruct a 3D shape of the deformed document, which inevitably brings lots of inconveniences so as to block their further real applications. 
Recently, learning-based methods~\cite{8578592,li2019document,xie2020dewarping,feng2021doctr,feng2022geometric,9010747} estimate a dense flow field from the distorted image to the distortion-free one, which have shown promising performance. With the flow field, the input distorted image can be unwarped for rectification. Although these methods report superior performance, how to learn effective representations of a distorted document image is still under-explored.   

\begin{figure}[t]
  \centering
    \includegraphics[width=1\linewidth]{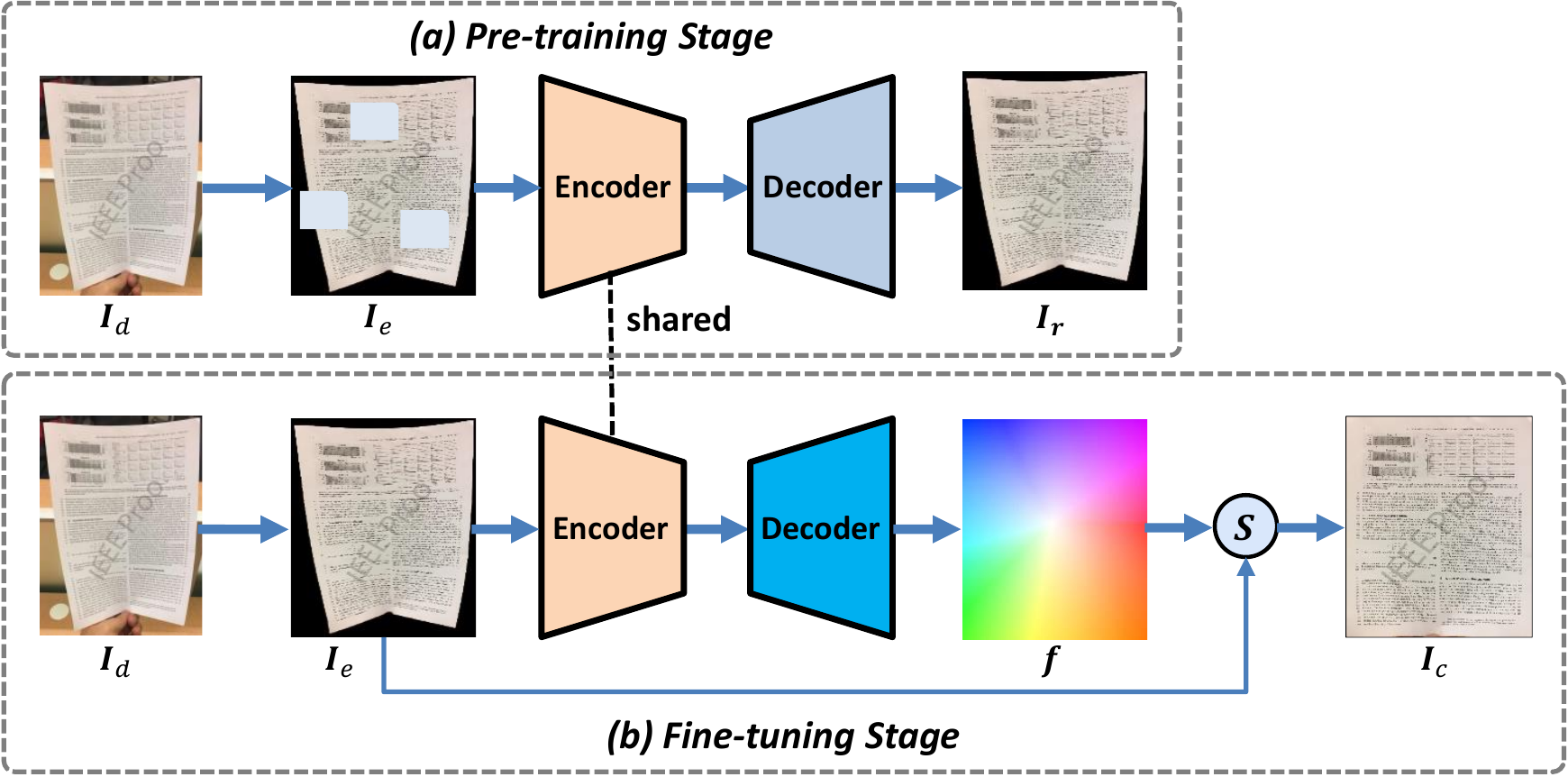}
    \caption{An overview of our DocMAE. 
    It consists of two stages: 
    (a) A pre-training stage that reconstructs the randomly masked patches;
    (b) A fine-tuning stage that transfers the learned representations for distortion rectification. ``$S$'' denotes the warping operation based on bilinear sampling.}
    \label{fig:overview}
\end{figure}

In document image rectification, it is crucial to extract the structural information of the deformed document. In a document image, the most informative cues for rectification exist in the document boundaries and text lines. Concretely, the document boundaries contain information about the global physical deformation and shooting angles, while the text lines show the deformation of local regions.
Besides, there is an explicit prior on the text lines that the distorted horizontal text lines are straight in the rectified image. To obtain the structure representation,
traditional methods resort to auxiliary hardware~\cite{937649,4407722,6909892,brown2007} or multi-view images
~\cite{4916075,you2017multiview,yamashita2004shape}.
%~\cite{brown2004image,4916075,yamashita2004shape,you2017multiview}.
Recently, learning-based method~\cite{9010747} have learnt a 3D coordinate map with a U-Net~\cite{ronneberger2015u}.
However, these existing structure representation learning methods for document images require auxiliary hardware or human supervision.

To avoid the inconvenience of existing structure representation methods, we propose DocMAE, a self-supervised learning-based framework for document image rectification, inspired by the success of MAE~\cite{he2022masked}.
The framework of our DocMAE is simple, which consists of a pre-training stage for distortion representation learning and a fine-tuning stage for distortion rectification. 
Technically, during the pre-training stage, we first mask the random patches of the background-excluded document images and then reconstruct the missing pixels.
Note that here the reconstruction is conducted on the background-excluded document images to avoid non-unique solutions because a document can be placed in various scenes.
Besides, to support pre-training process, we collect a large-scale synthetic document distortion dataset named LDIR, which fully simulates the various distortion of real document images.
Then, in fine-tuning stage, the learned representations are transferred to the downstream rectification task.

% Extensive experiments are conducted on the challenging Doc3D dataset~\cite{9010747} and DocUNet benchmark dataset~\cite{8578592}.
Extensive experiments are conducted on our proposed LDIR dataset, Doc3D dataset~\cite{9010747}, and DocUNet benchmark dataset~\cite{8578592}.
The results verify the effectiveness of our method as well as its superior performance over existing methods.
In summary, we make three-fold contributions as follows:
\begin{itemize}
    \item 
    We propose DocMAE, a self-supervised learning-based framework for document image rectification.
    \item 
    We propose a large-scale dataset based on the rendering techniques for self-supervised representation learning.
    \item
    We conduct extensive experiments to verify the merits of our method and report the state-of-the-art performance.
\end{itemize}

\section{Related Work}

There are mainly two technical routes to address document image
rectification, including (a) rectification based on 3D reconstruction, and (b) rectification based on low-level features. We discuss them separately in the following.

\subsection{Rectification Based on 3D Reconstruction}
In order to rectify document images, some traditional methods take advantage of auxiliary equipment to reconstruct the 3D shape of the deformed documents and then flatten the reconstructed surface to correct the distortions. Brown and Seales~\cite{937649} utilize a light projector to gain the 3D representation of the document shape and then flatten the page through a spring-mass particle system. Zhang~et al.~\cite{4407722} fulfill restoration based on physical modeling techniques with the help of a laser scanner. Meng~et al.~\cite{6909892} project two structured beams illuminating the document page to recover two spatial curves of the page surface. In comparison, some other methods exploit multi-view images for 3D shape reconstruction. Among them, Koo~et al.~\cite{4916075} calculate the corresponding points between two document images by SIFT to resolve the unfolded surface. You~et al.~\cite{you2017multiview} present a method based on a ridge-aware 3D reconstruction technique. Tan~et al.~\cite{tan2005restoring} employ the shading technique to acquire shape for distortion rectification. Das~et al.~\cite{das2017common} build a 3D shape model with the help of the convolutional neural network. However, no matter whether using auxiliary equipment or taking advantage of multi-view images, these methods cannot be used in common situations, resulting in the limitation of their usability.

\subsection{Rectification Based on Low-level Features}
The low-level features of an image also contain informative cues for geometric distortion rectification. In previous work, many algorithms focus on how to restore the curved text lines straight. For example, Lavialle~et al.~\cite{958227} model the detected text lines as cubic B-splines. While Mischke and Luther~et al.~\cite{mischke2005document} utilize polynomial approximation to model it. However, these methods rely more on hand-craft settings and human prior knowledge. The neural network is introduced to solve this task by Ma~et al.~\cite{8578592}. They utilize a stacked UNet to directly regress the pixel-wise displacement. Li~et al.~\cite{li2019document} stitch the displacement field of the image patches to unwrap the image. Xie~et al.~\cite{xie2020dewarping} adopt a fully convolutional network to evaluate pixel-wise displacement. FDRNet~\cite{xue2022fourier} transforms the image to the Fourier space to extract structural representations. 
% More recently, Feng~et al.~\cite{feng2021doctr} first introduce transformer as a stronger backbone to solve rectification tasks. 
% In \cite{feng2022geometric}, Feng~et al. train a hybrid transformer network, which combines 3D shape and text lines information to correct image distortion. 

\section{APPROACH}

In this section,
we present DocMAE, a novel framework for the geometric rectification of document images.
Fig.~\ref{fig:overview} shows the framework of our method.
DocMAE consists of two stages, including: (1) a distortion-aware pre-training stage that reconstructs the randomly masked patches, and (2) a rectification fine-tuning stage that transfers the learned representations for estimating the distortion rectification.

% Based on the task-specific self-supervised pre-training,
% the proposed DocMAE performs the representation learning of input images,
% including the structural properties of the geometric distorted images and 
% illumination distribution of the geometric rectified images.
% After that, we transfer the learned representation for the fine-tuning of different rectification tasks.

\subsection{Self-supervised Pre-training}
In a geometrically distorted document image, 
the document structure information is reflected by its edges, text lines, and illumination variations,
which provides rich cues for distortion rectification.
To obtain the structure representation in a convenient way,
our DocMAE framework introduces a distortion-aware self-supervised pre-training stage, free of hardware requirements or human supervision.

\smallskip
\noindent
\textbf{Background Removal.} 
% Due to the diversity of document image background, the processing of reconstructing the background cannot help the network to learn the structure information of the document, which is different from reconstructing text lines and paper edges. Therefore, in order to obtain the meaningful feature of the document image, the background of the image needs to be removed during each stage. In the pre-training stage, we utilize the synthetic dataset LDIR constructed by ourselves to avoid this issue. Then, a lightweight semantic segmentation network is used to remove the image background during the fine-tuning stage or evaluation stage. This step also alleviates the non-unique problem because the document can be placed anywhere.
Due to the diversity of the document image background, the reconstruction of them cannot help the learning of the structure information, different from the reconstruction of text lines and document boundaries. Therefore, to obtain the meaningful features for rectification, we remove the background of the input image $\boldsymbol{I}_d \in \mathbb{R}^{H \times W \times 3}$ during the pre-training stage. Specifically, a lightweight semantic segmentation network~\cite{qin2020u2} is trained to predict the mask $\bm{M} \in \mathbb{R}^{H \times W \times 1}$ of the foreground document.
Then, the noisy background is removed by pixel-wise multiplication along the channel dimension between image $\boldsymbol{I}_d$ and mask $\boldsymbol{M}$.
% This step also alleviates the non-unique problem because the document can be placed anywhere. Then, we gain background-excluded image $\boldsymbol{I}_e$.

% In order to obtain the meaning feature of the document image and resolve the non-unique problem, the background of these images needs to be removed. In the training stage, thanks to the convince of the synthetic dataset LDIR, which was built by ourselves. We can utilize the world coordinate to manipulate and gain the background-excluded image based on the original distorted document image. As for the testing and evaluation stage, we build up an in-front light semantic segmentation network to predict the confidence matrix of the foreground document. The confidence matrix will be filtered with a certain threshold and then become a barbarized mask. After that, the background of the testing image can be removed through matrix multiplication with the mask by broadcasting. This background-removing module can also be used in further real-world inference processing.

\begin{figure*}[t]
  \centering
%   \fbox{\rule{0pt}{2in} \rule{0.9\linewidth}{0pt}}
    \includegraphics[width=1\linewidth]{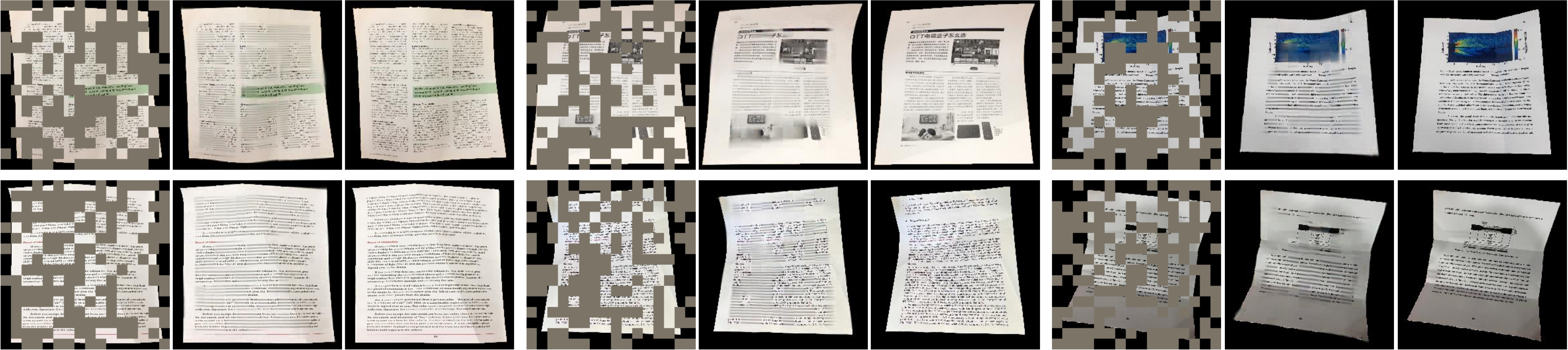}
    % \vspace{-0.06in}
    \caption{Example results of real document images. For each triplet, we show the mask image (left), our reconstruction (middle), and the ground truth (right). The masking ratio is set as 50$\%$ here.}
    
  \label{fig:mae}
% \vspace{-0.1in}
\end{figure*}

% \smallskip
% \noindent
% \textbf{Pre-training via Reconstruction. } After the background removal step, we devise a self-supervised pre-training module to acquire the domain knowledge of document images. Specifically, we utilize our collected synthetic dataset LDIR, including twenty thousand distortion document images, for self-supervised training. As shown in Fig.~\ref{fig:overview}, given a geometrically distorted document image, we randomly mask patches of the input image and reconstruct the missing pixels with Masked Auto-Encoders (MAE)\cite{he2022masked}.
% Note that the original MAE does self-supervised pre-training on the large-scale ImageNet-1K dataset~\cite{deng2009large}, where objects are of various categories complex under complex scenes. 
% In our pre-training stage, we utilize MAE to learn the structure representation in document image
% rectification, which is a largely different low-level vision task. The network first needs to complete the missing edges.
% Besides, since for geometric rectification, the input is a low-resolution image (often 288 $\times$ 288),
% we just need to reconstruct the coarse-grained text lines based on the context from visible patches. Furthermore, the network also should maintain the continuity of the illumination.
% Hence, this is a nontrivial and meaningful self-supervisory task.

\smallskip
\noindent
\textbf{Masking.}
% To process background-excluded document image $\boldsymbol{I}_e \in \mathbb{R}^{H \times W \times 3}$, 
% Following .. ~\cite{vit},
Following ViT~\cite{ViT}, we first divide the background-excluded document image $\boldsymbol{I}_e \in \mathbb{R}^{H \times W \times 3}$ into a sequence of 2D patches $\boldsymbol{x}_p \in \mathbb{R}^{N \times\left(P^2 \cdot 3\right)}$, where $H$ and $W$ represent the height and width of the document image $\boldsymbol{I}_d$, $P$ represents the patch size, and $N=H W / P^2$ denotes the number of patches. Then, we randomly mask the $N$ patches $\boldsymbol{x}_{p}$ with a ratio $R$.

\smallskip
\noindent
\textbf{Distortion Encoder.}
The distortion encoder extracts the features of the input image $\bm{I}_e$.
We only process the visible patches $\boldsymbol{x}_{v} \in \mathbb{R}^{N_v \times\left(P^2 \cdot 3\right)} $, where $N_v = N \times (1-R)$ denotes the number of patches. Then, these patches are flattened and mapped to $D$ dimension with a linear projection. The output is the patch embeddings $\boldsymbol{E}_o \in \mathbb{R}^{N_v \times D}$. Here, we set $D=512$. To maintain the positional information, positional embeddings  $\boldsymbol{E}_p \in \mathbb{R}^{N_v \times D}$ (the sine-cosine version) are included and bonded with the patch embeddings $\boldsymbol{E}_o$. Then, the output passes through $K_1$ transformer blocks~\cite{ViT} to output encoded visible patches $\boldsymbol{E}_e \in \mathbb{R}^{N_v \times D}$.

\smallskip
\noindent
\textbf{Reconstruction Decoder.}
The learnable mask tokens $\boldsymbol{E}_{m} \in \mathbb{R}^{N_m \times D} $ are zero-initialized and concatenated with encoded visible patches $\boldsymbol{E}_e$, where $N_m = N \times R$ is the number of masked patches. Then we add positional embeddings $\boldsymbol{E}_p' \in \mathbb{R}^{N \times D}$ to all tokens, to help mask tokens gain the information about their locations in the image. The obtained embeddings $\boldsymbol{E}_d \in \mathbb{R}^{N \times D}$ are then fed into another $K_2$ transformer blocks~\cite{ViT}. Finally, we employ a linear projection to project the output channels $D$ to the pixel number in each patch, \emph{i.e.}, $P^2 \times 3$. The output $\boldsymbol{x}_f \in \mathbb{R}^{N \times\left(P^2 \cdot 3\right)}$ is reshaped to form the reconstructed image $\boldsymbol{I}_{r}  \in \mathbb{R}^{H \times W \times 3}$.

\smallskip
\noindent
\textbf{Loss Function.} 
% To learn a useful distortion representation for rectification, we introduce a pre-training loss $\mathcal{L}_{pre}$ to supervise the output image  $\boldsymbol{I}_{r}$. $\mathcal{L}_{pre}$ is calculated by comparing the reconstructed images $\boldsymbol{I}_{r}$ and background-excluded image $\boldsymbol{I}_{e}$ in the pixel space with the following objective. It should be noticed that we only compare the masked part,
The training loss is defined as the $L_2$ distance between the reconstructed image $\boldsymbol{I}_{r}$ and the input background-excluded image $\boldsymbol{I}_{e}$ on the masked patches,
\begin{equation}
\mathcal{L}_{pre} =\frac{1}{n} \sum_{i=1}^n\left(y_i-\hat{y}_i\right)^2,
\end{equation}
where $y_i$ denotes the pixel value of the background-excluded image $\boldsymbol{I}_{e}$, $\hat{y}_i$ represents the pixel value of the reconstructed image $\boldsymbol{I}_{r}$, and $n$ denotes the number of reconstructed pixels.

% We manipulate MSE as the loss of the network, which presents the similarity between the construction images and original images. Among the training of the network, the MSE is expected to become less. At the same time, the images will be reconstructed and become similar to the original image. In this processing, the network gain prior knowledge of the document, such as the document boundaries, text lines, and paragraph layout, which enhances the fine-tuning performance in the downstream task,e.g. document image rectification.

\subsection{Fine-tuning for Rectification}
In this section, we transfer the learned representations for downstream distortion rectification.
As shown in Fig.~\ref{fig:overview}, a rectification decoder is cascaded to the pre-trained encoder and outputs the rectified image, described next. 

% Thanks to the prior knowledge through self-supervised learning, the network will converge more rapidly, meanwhile, the rectification performance will be better than the model without any pre-training.  

% \smallskip
% \noindent
% \textbf{Flow Prediction. }

\smallskip
\noindent
\textbf{Feature Extraction.}
% In order to improve the accuracy of flow prediction, feature extraction is utilized. 
% After the self-supervised learning is fulfilled, the network is familiar with the features of distorted document images, \textit{e.g.}, text lines, structural patterns, and lighting conditions. 
Given an input image $\boldsymbol{I}_{d} \in \mathbb{R}^{H \times W \times 3}$, we first remove its noisy background as the pre-training stage, to make the rectification network focus on the distortion rectification without localizing the document first.
Then, the obtained background-excluded document image $\boldsymbol{I}_e \in \mathbb{R}^{H \times W \times 3}$ is divided into multiple patches, embedded into tokens, and fed into the pre-trained distortion encoder for representation extraction.
Next, we fed the obtained representations into another rectification decoder with $K_2$ transformer blocks~\cite{ViT}.
The output feature $\bm{E}_f \in \mathbb{R}^{N \times D}$ is taken as the input of the following flow prediction head.
% The weights of the encoder in the pre-training network will be shared with the encoder in the fine-tuning network, which can boost the performance of distortion image rectification. As for the decoder in the pre-training network, since it contains more knowledge about masking pixel reconstruction, its weights will be dropped and filtered out in the fine-tuning stage. The image features $\boldsymbol{F}_{d}$ are extracted through the network.

% We only initialize the first $N_F$ ($N_F$ < $N_T$ ) layers with pre-trained weights, and leave the last layers randomly initialized.

\smallskip
\noindent
\textbf{Flow Prediction.}
We first reshape the feature $\bm{E}_f \in \mathbb{R}^{N \times D}$ to $\bm{f}_s \in \mathbb{R}^{\frac{H}{P} \times \frac{W}{P} \times 2}$.
Then, we upsample the $\frac{1}{P}$ scale warping flow $\bm{f}_s$ to full-scale one $\boldsymbol{f}\in \mathbb{R}^{H \times W \times 2}$ using a learnable upsample module~\cite{feng2021doctr,feng2021docscanner}. Note that $\boldsymbol{f}$ is a flow field that describes the deformation from the distorted image to the distortion-free one. 
 Given the predicted warping flow $\boldsymbol{f} \in \mathbb{R}^{H \times W \times 2}$, we resample the pixels from the background-excluded image $\boldsymbol{I}_{e}$ to generate the rectified one $\boldsymbol{I}_{c} \in \mathbb{R}^{H \times W \times 3}$ as follows, 
\begin{equation}\label{equ:task}
	\bm{I}_c(u,v) = \bm{I}_e(\bm{f}_u(u,v), \bm{f}_v(u,v)),
\end{equation}
where $(u,v)$ is the integer pixel coordinate in rectified image, $\bm{f}_u$ and $\bm{f}_v$ represent the two channel of warping flow $\bm{f}$, and $(\bm{f}_u(u,v), \bm{f}_v(u,v))$ is the projected coordinate in $\bm{I}_e$.

\smallskip
\noindent
\textbf{Loss Function.}
% In order to improve the performance of document image rectification, a fine-tuning loss $\mathcal{L}_{ft}$ is introduced to enhance the accuracy of flow prediction,
During the fine-tuning stage, the model is optimized with the training objective as follows,
\begin{equation}
\mathcal{L}_{ft} =\frac{1}{n} \sum_{i=1}^n\left|y_i-y'_i\right|,
\end{equation}
where $y_i$ and $y'_i$ are the coordinates in predicted flow $\bm{f}$ and ground truth $\bm{f}_{gt}$, and $n$ denotes the number of pixels in $\bm{f}$.
% and $\mathcal{L}_{ft}$ is the average $L_1$ distance between the predicted warping flow $\boldsymbol{f}$ and its given ground truth $\boldsymbol{f}_{gt}$. %$\in \mathbb{R}^{H \times W \times 3}$.

% $\mathcal{L}_{ft}$ is used to represent the difference between the rectification image and the ground truth image. We calculate the absolute error between each pixel. After that, we compute the average value of it. We utilize $\mathcal{L}_{ft}$  as the loss of this network, which presents the distance between the predicted backward mapping  $\boldsymbol{f}_{b}$  with ground truth flow,
% \begin{equation}
% \mathcal{L}_{ft} =\frac{1}{n} \sum_{i=1}^n\left|y_i-y'_i\right|,
% \end{equation}
% where $n$ denotes the total number of pixels in each image, $y_i$ represents the pixel value of the ground truth flow $\boldsymbol{f}_{gt}$, and $y'_i$ denotes backward mapping $\boldsymbol{f}_{b}$.

 % Among the training of this network, the MAE is expected to become less, and the images will be restored and become similar to the ground truth image. In this processing, the network learns how to correct distorted document images with the help of prior knowledge and more specific feature gained in the fine-tuning stage.

\section{Experiment}

% As for self-supervised learning, We use our self-generated dataset. The task is randomly masking patches of document images from the dataset and training the network to reconstruct the missing pixels. Then we do supervised learning on the Doc3D~\cite{9010747} dataset to train the transformer for geometric unwrapping tasks. 

\subsection{Dataset}

% In order to fulfill document image rectification, a large amount of data is required to support deep learning. However, the datasets in this field are scarce. 

% \smallskip
\noindent
\textbf{Doc3D.}
% As a popular evaluation dataset, 
Doc3D dataset~\cite{9010747} consists of 100k distorted document images created by real document data with rendering software. 
It is the largest dataset to date in the field. 
In this work, we take the Doc3D dataset for training in our approach. 

\smallskip
\noindent
\textbf{LDIR.}
To perform self-supervised learning on document images, we propose LDIR, a large-scale synthetic dataset for document image rectification. 
We utilize 3D rendering software to simulate real-world document textures, lighting conditions, backgrounds, and distortions, which ensures the quality of LDIR.
It contains 200k distorted document images. The distorted document images are rendered through real document data in our daily life, such as books and magazines. 
Our experiments reveal the high quality of LDIR that indeed improves the rectification performance based on self-supervised learning.

\smallskip
\noindent
\textbf{DocUNet benchmark.} 
DocUNet benchmark dataset~\cite{8578592} is a widely-used dataset for the evaluation of rectification algorithms. It contains 130 real-world distorted document images and their scanned ground truth. 
% However, the datasets in this field are scarce. Moreover, the majority of existing datasets cannot simulate the real distorted document image, which lacks lighting conditions, texture simulation, and curve degree variation. 

\subsection{Setup}

We use all the images of LDIR dataset for pre-training and all the images of Doc3D dataset~\cite{9010747} for fine-tuning. The image size $(H, W)$ is (288,288) and the patch size $P$ is 16. During the pre-training stage and fine-tuning stage, the latent vector size $D$ of the encoder and decoder is both 512. 
% The total layer number $N_t$ is both 10.
The layer number $K_1$ and $K_2$ are set as 6 and 4, respectively.

We use the Adam
%~\cite{kingma2014adam} 
optimizer and one-cycle learning rate policy with a maximum value of $10^{-4}$. Both stages are trained for 65 epochs with a batch size of 64. Two NVIDIA GeForce RTX
1080Ti GPUs are employed to train the network.

\subsection{Evaluation Metrics}

% As for geometric unwrapping, 
We discuss the evaluation scheme mainly in two fields: pixel alignment and Optical Character Recognition (OCR) accuracy. 
% In detail, 
Specifically,
for pixel
alignment, Local Distortion (LD)
%~\cite{you2017multiview} 
and Multi-Scale Structural SIMilarity (MS-SSIM)
%~\cite{1292216} 
are recommended to evaluate the restoration performance as previous works~\cite{8578592,xie2020dewarping,9010747} suggest. 
In terms of OCR, Edit Distance (ED)
%~\cite{levenshtein1966binary} 
and Character Error Rate (CER)
%~\cite{morris2004and} 
are used to evaluate the performance on text recognition, following~\cite{8578592,9010747}.

\smallskip
\noindent
\textbf{Local Distortion.} 
Local Distortion (LD)~\cite{you2017multiview} calculates the average deformation of each pixel and represents the mean displacement error according to the SIFT flow $(\Delta x, \Delta y)$~\cite{5551153} from the ground-truth scanned image to the rectified one. 
% The SIFT flow~\cite{5551153} demonstrates a 2D map relationship through each pixel from the scanned image toward
% the rectified image. 
% In conclusion, LD is equal to the $L_2$ distance between all matched pixels. 

\smallskip
\noindent
\textbf{MS-SSIM.} 
% The Structural SIMilarity (SSIM)~\cite{1284395} represents the similarity between two images. 
% It is manipulated on various windows of an image. 
% The sampling density of an image effect the calculation quality significantly. Therefore,
The Multi-Scale Structural SIMilarity (MS-SSIM) calculates the multi-scale image similarity
%~\cite{1284395} 
between the ground-truth scanned image to the rectified one.
We follow the weights setting of works~\cite{8578592,xie2020dewarping,9010747}.
% In addition, all of the weights of SSIM among multiple resolutions will be summed up into MS-SSIM~\cite{1292216}.

\smallskip
\noindent
\textbf{ED and CER.} Edit Distance (ED)
%~\cite{levenshtein1966binary} 
measures the differences between two strings, based on the minimal number of operations required to change one string into the target one. It involves three types of operations, including deletions ($d$), insertions ($i$), and substitutions ($s$). Furthermore, Character Error Rate (CER) can be computed: $(d+i+s) / N_s$, where $N_s$ is the total number of the target string. 
% We choose PyTesseract~\cite{4376991} as the OCR engine to evaluate the accuracy, following ~\cite{9010747,8578592}.

\subsection{Comparison with State-of-the-art Methods}

We evaluate the performance of DocMAE on the DocUNet Benchmark dataset~\cite{8578592} by quantitative and qualitative evaluation. Table~\ref{com:b1} shows the comparisons of our DocMAE with the existing state-of-the-art learning-based methods~\cite{8578592,li2019document,xie2020dewarping,9010747,das2021end,xie2021document,xue2022fourier} on distortion metrics, including distortion rectification and OCR accuracy. 

% Note that for OCR accuracy evaluation, following
% DewarpNet~\cite{9010747} and DocTr~\cite{feng2021doctr}, we select 50 and 60 images
% from the DocUNet Benchmark dataset~\cite{8578592}, respectively,
% where the text makes up the majority of content. This is
% because if the text is rare in an image, the character number
% Nc (numerator) in Equation is a small number, leading
% a large variance for CER.

% \smallskip
% \noindent
% \textbf{Comparison with state-of-the-art methods.} 
% As shown in Table~\ref{com:b1}, DocMAE achieves superior performance on LD metric compared with other methods. As for ED, CER, and MS-SSIM metrics, DocMAE reaches an average level compared with other methods. 

\begin{figure*}[t]
  \centering
    \includegraphics[width=0.99\linewidth]{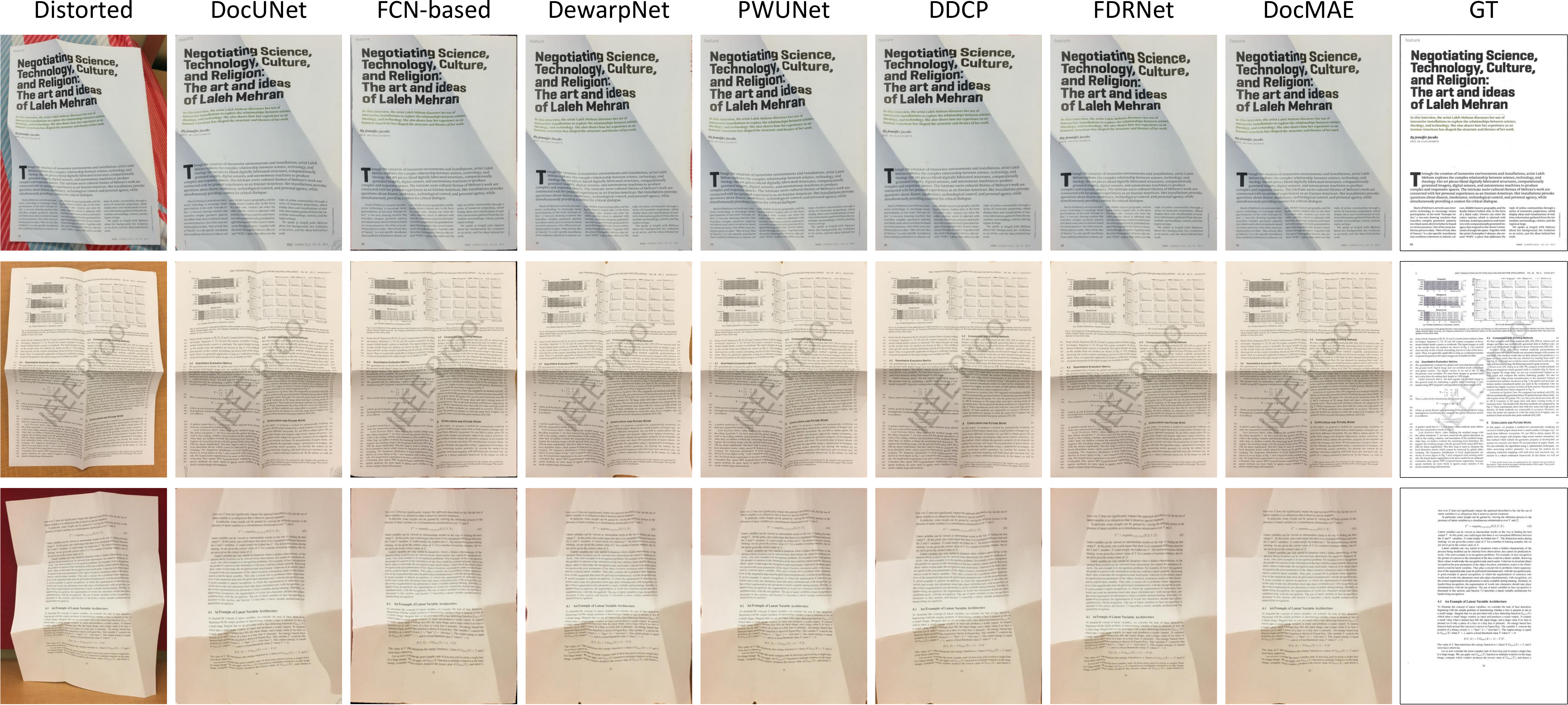}
    % \vspace{-0.05in}
    \caption{Qualitative comparison on DocUNet benchmark dataset~\cite{8578592}. Note that such images are real-world document images. Our DocMAE can effectively rectify such images and show less distortion compared with other learning-based methods.}
  \label{fig:compare}
  % \vspace{-0.13in}
\end{figure*}

As it can be seen in Fig.~\ref{fig:compare}, DocMAE achieves excellent qualitative rectification.  Compared with other learning-based methods, the rectified images of DocMAE show less distortion and the corrected text lines are more straight. 

\setlength{\tabcolsep}{0.2mm}
\begin{table}[t]
  	\small
	\caption{Quantitative comparisons of the existing learning-based methods in terms of distortion metrics, OCR accuracy, and image similarity on the DocUNet Benchmark dataset.
 % and running efficiency on the DocUNet Benchmark dataset.
	% ``*'' denotes that the OCR metrics could not be calculated as the rectified images or models are not publicly available.
    ``$\uparrow$'' indicates the higher the better, while ``$\downarrow$'' means the opposite.}
 	% \vspace{0.05in}
	\centering
	
	\begin{tabular}{c|c|c|cc|c}  
		\Xhline{0.2\arrayrulewidth}
		Methods  & Venue &LD $\downarrow$ & ED $\downarrow$ &CER $\downarrow$ &MS-SSIM $\uparrow$   \\  
		
		\hline\hline
		
		Distorted Images & - & 20.51 &2111.56 & 0.54 & 0.25   \\ 
        \hdashline
		DocUNet~\cite{8578592} & \emph{CVPR'18} & 14.19 & 1933.66 & 0.46 & 0.41     \\
		% AGUN* & \emph{PR'18} & 12.06 & - & - & 0.45   \\
		DocProj~\cite{li2019document} & \emph{TOG'19} & 18.01 & 1712.48 & 0.43 & 0.29   \\ 
		%DocUNet on Doc3D~\cite{9010747} & 0.4389 & 10.90 & - & 1145.97 & 0.3816 & - & - \\   
		DewarpNet~\cite{9010747} &  \emph{ICCV'19} & 8.39 & 885.90 & 0.24 & 0.47  \\  
 		FCN-based~\cite{xie2020dewarping} & \emph{DAS'20} & 7.84 & 1792.60 & 0.42 & 0.45   \\
% 		DocTr~\cite{feng2021doctr} &  \emph{ACM MM'21} & 8.38 & 839.82 & 0.27 & \textbf{0.51} & 7.40 & 26.9 \\ 
		PWUNet~\cite{das2021end} &  \emph{ICCV'21} & 8.64 & 1069.28 & 0.27 & 0.49  \\  
		%\Xhline{1\arrayrulewidth}
		% DocTr~\cite{feng2021doctr} &  \emph{MM'21} & 7.76 & 724.84 & 0.18 & 0.51  \\ 
		DDCP~\cite{xie2021document} &  \emph{ICDAR'21} & 8.99 & 1442.84 & 0.36 & 0.47   \\ 

            FDRNet~\cite{xue2022fourier}  &  \emph{CVPR'22} & 8.21 & 829.78 & 0.21  & \textbf{0.54}   \\ 

  %             RDGR~\cite{jiang2022revisiting}  &  \emph{CVPR'22} & 8.51 & 729.52 & \textbf{0.17} & 0.49   \\ 
		% DocGeoNet~\cite{feng2022geometric} &   \emph{ECCV'22} & 7.71 & \textbf{713.94} & 0.18 & 0.50    \\ 
% 		\hline
  
		\hline
		DocMAE (Ours) & - & \textbf{7.63} & \textbf{801.52}  & \textbf{0.20} & 0.50   \\ 
		\Xhline{0.2\arrayrulewidth}	
	\end{tabular}
  % \vspace{-0.04in}
	\label{com:b1}
\end{table}

\subsection{Ablation Studies}
In this section, we conduct ablations to verify the effectiveness of each component in DocMAE, including the self-supervised pre-training, the masking strategy, the way of fine-tuning, and the dataset for pre-training.

% \subsubsection{Self-supervised pre-training.}

% The key idea of our work is introducing self-supervised pre-training into the document image rectification field. Table 2 ~\ref{com:b2} compare the performance between the baseline network and fine-tuning network. It is obvious that pre-training help the DocMAE gain significant improvement when compared with the baseline network without any pre-training. The fine-tuning network surpasses the baseline network in every metric, which proves that self-supervised learning  can help the network gain more prior knowledge, such as text lines, lighting condition, and paragraph layout during the pre-training stage, resulting in the improvement in the downstream document image restorations. 

\smallskip
\noindent
\textbf{Self-supervised Pre-training.}
The key idea of DocMAE is the self-supervised representation learning strategy for document images. We study the impact of self-supervised learning strategy on the learned representations in Table~\ref{com:b2}. 
As we can see, self-supervised representation learning significantly boosts the rectification performance.
This can be attributed to the representation learning
of the structural cues in document images to improve the rectification.
Furthermore, as shown in Fig.~\ref{fig:mae}, we show some cases of the reconstructed results using the pre-trained model. 
The document boundaries and text lines are well-reconstructed. Note that here our goal is not to reconstruct the fine-grained text lines, but to capture their layout caused by perspective and paper distortions.

\setlength{\tabcolsep}{0.9mm}
\begin{table}[t]
   \normalsize
 % \caption{Comparison of the performance between the Baseline network and the fine-tuning network. The Baseline network is a pure document image restoration network. The fine-tuning network extract features from previous self-supervised learning.  }
  \caption{Ablations experiments about the pre-training stage for representation learning. With the learned representations, the rectification performance improves significantly.}
 	% \vspace{0.05in}
 \centering
 
 \begin{tabular}{c|c|cc|c}  
  \Xhline{0.7\arrayrulewidth}
  Methods   &LD $\downarrow$ & ED $\downarrow$ &CER $\downarrow$ &MS-SSIM $\uparrow$    \\  
  
    \hline 
  w/o pre-training &  8.69 & 854.84 & 0.23  & 0.48  \\ 
  % \hline
  w/ pre-training    & \textbf{7.63} & \textbf{801.52}  & \textbf{0.20} & \textbf{0.50}   \\ 
  \Xhline{0.7\arrayrulewidth} 
 \end{tabular}
  % \vspace{-0.1in}
 \label{com:b2}
\end{table}

% \subsubsection{Masking ratio.}
\smallskip
\noindent
\textbf{Masking Ratio.}
The masking ratio affects the difficulty of self-supervised learning. Therefore, we initialize the pre-training network with different mask ratios. As the result shown in Table~\ref{com:b3}, DocMAE achieves the best performance with a 75$\%$ mask ratio. With a higher mask ratio, more details of the document like edges and text lines are missing, which makes the network struggle to extract high-level structural representations. In contrast, with a lower mask ratio, the task is relatively easy for the network to learn effective representations.
% distinguish more detailed and valuable information.

\setlength{\tabcolsep}{0.8mm}
% change h to ht
\begin{table}[t]
   \normalsize
 % \caption{ Ablation experiments about masking ratio. As for self-supervised learning, choosing a moderate-difficulty task is crucial. Several masking ratios of the image have been tested and recorded. The results are listed in the following Table~\ref{com:b3}. When the mask ratio comes to 75\%, the network achieves the best performance.}
 \caption{Ablation experiments about masking ratio in the pre-training stage. $75\%$ produces the best performance.}
  % \vspace{0.05in}
 \centering
 
 \begin{tabular}{c|c|cc|c}  
  \Xhline{0.7\arrayrulewidth}
  Masking Ratio   &LD $\downarrow$ & ED $\downarrow$ &CER $\downarrow$ &MS-SSIM $\uparrow$  \\  
    \hline 
    
  %     25\%  &  8.15 & 863.06 & 0.22 & \textbf{0.50}   \\ 
  % % \hline 
  % 50\% &   8.03 & \textbf{709.44} & \textbf{0.18} & 0.49  \\ 
  % \hline   
        %     15\%  & - & - & - & \textbf{-}   \\ 

      %     30%  &  - & - & - & -   \\ 

    %     45\%  &  - &- & - & -   \\ 
    
    60\%   & 7.88 &  808.35 & \textbf{0.20} & \textbf{0.50}   \\ 
  % \hline   
  75\%    & \textbf{7.63} & \textbf{801.52}  &  \textbf{0.20}  & \textbf{0.50}   \\ 
 
   % \hline   
    % 80\%  &  running & - & - & -   \\ 
    %    \hline  
  90\%  &  8.21 & 933.52 & 0.22 & 0.49   \\ 
  
  \Xhline{0.7\arrayrulewidth} 
 \end{tabular}
  % \vspace{-0.1in}
 \label{com:b3}
\end{table}

\setlength{\tabcolsep}{0.05mm}
\begin{table}[t]
   \normalsize
 % \caption{ Performance comparison between the network fixed the pre-training weights and unfixed these weights. As the result shows, in the fine-tuning stage, it is also necessary to train the weights inherited from the pre-training.  }
 \caption{Ablation experiments about fine-tuning way. Fine-tuning the whole model works better.}
 	 % \vspace{0.05in}
 \centering
 
 \begin{tabular}{c|c|cc|c}  
  \Xhline{0.7\arrayrulewidth}
  Settings   &LD $\downarrow$ & ED $\downarrow$ &CER $\downarrow$ &MS-SSIM $\uparrow$     \\  
  
    \hline 
    
  %  Fixed fine-tuning &   8.56 & - & -  & 0.49   \\ 
  % \hline
  %   Normal fine-tuning  & 7.53 & 803.160 & 0.1995 & 0.49   \\ 

   freezing the encoder  &   8.56 & 1011.64 & 0.25 & 0.49   \\ 
  % \hline
    fine-tuning the whole model & \textbf{7.63} & \textbf{801.52}  & \textbf{0.20} & \textbf{0.50}   \\ 
    
  \Xhline{0.7\arrayrulewidth} 
 \end{tabular}
 % \vspace{-0.1in}
 \label{com:b4}
\end{table}

\setlength{\tabcolsep}{1.2mm}
\begin{table}[t]
   \normalsize
 \caption{Ablation experiments about the dataset used for pre-training.
 The LDIR dataset helps gain more prior knowledge compared with the Doc3D dataset~\cite{9010747}.
 }
 	% \vspace{0.05in}
 \centering
 
 \begin{tabular}{c|c|cc|c}  
  \Xhline{0.7\arrayrulewidth}
  Dataset  &LD $\downarrow$ & ED $\downarrow$ &CER $\downarrow$ &MS-SSIM $\uparrow$   \\  
  
    \hline 
    
   Doc3D &  8.05 & 891.76 & 0.22  & 0.49  \\ 
  % \hline
    LDIR  & \textbf{7.63} & \textbf{801.52}  & \textbf{0.20} & \textbf{0.50}   \\ 
  \Xhline{0.7\arrayrulewidth} 
 \end{tabular}
  % \vspace{-0.1in}
 \label{com:b5}
\end{table}

 % \subsubsection{Fixed weights.}
\smallskip
\noindent
\textbf{Fine-tuning Way.}
Table~\ref{com:b4} studies the impact of fine-tuning way on performance.
By default, during the fine-tuning stage, we update the weights of the whole model for rectification.
% During the fine-tuning stage, normally, the feature weights inherited from the pre-training stage are also trained. 
Then, we fixed the pre-trained encoder and only fine-tune the rectification decoder.
As we can see, our default fine-tuning way produces much better performance.
% weights to observe the difference. Obviously, compared with freezing the encoder, it is better to fine-tune the whole model.

% Although self-supervised learning promotes the network to achieve novel performance, the difference between the change of the weights of the pre-training model is also a critical question. Therefore, we establish a contrast experiment to explore this question. In one experiment, we load the weights of the pre-training model and also let them change during the fine-tuning processing in terms of varying their gradient value. In the other experiment, we also import all of the weights of the pre-training model but fix them during the fine-tuning processing. After these two experiments finished, we compared all the matrices shown in Table~\ref{com:b4}. As Table~\ref{com:b4}shows, the network with all parameter variables performs better than the network fixed with the initially imported variable. It indicated that although pre-training provides the network with aware prior knowledge, the network still necessarily learns more detailed information during the fine-tuning stage, which indeed helps improve the document image restoration and overcome the over-fitting issue.

 % \subsubsection{ pre-training Datasets.}
\smallskip
\noindent
\textbf{Datasets for Pre-training.}
We ablate the dataset used for pre-training. We use two different datasets separately: the Doc3D dataset~\cite{9010747} and our LDIR dataset. As shown in Table~\ref{com:b5}, the network pre-trained on our LDIR dataset performs much better.
We attribute this performance gain to the use of the extra data and the quality of our LDIR dataset.
% Due to the large volume of the LDIR dataset, it provides the network with more details.
% Moreover, it is recommended to do self-supervised pre-training on a large-scale dataset, like the LDIR dataset, and do fine-tuning on a more specific dataset, such as the Doc3D~\cite{9010747} dataset. 

\section{CONCLUSION}

In this work, we present DocMAE, a self-supervised framework for document image rectification.
The key idea is to capture the structural cues in document images and leverage it for rectification.
Technically, we first mask random patches of the background-excluded document images and then reconstruct the missing pixels. 
% We fine-tune model with the encoder inherited from pre-training stage for document image rectification.
% Transfer performance validates the effectiveness of the learned representations.
In our implementation, we collect a large-scale dataset named LDIR based on the rendering techniques.
Extensive experiments are conducted, and the results demonstrate the effectiveness of the learned representations as well as the superior rectification performance.
% It contains two sub-network: the pre-training network and the fine-tuning network. According to the ablation studies, the model gains significant improvement through self-supervised learning, compared with the model without any pre-training. We also propose a new synthetic dataset containing two hundred thousand
% distortion images, called LDIR. The novel design makes DocMAE achieve the state-of-the-art performance and proves self-supervised learning can indeed enhance performance in document image rectification. In the future, we will explore more suitable tasks for improving self-supervised learning and shift this architecture also in illumination correction tasks. Besides, the LDIR will be enriched with multi-language distortion document images to enhance the model generalization ability. We will seek the solution in our future research.

\bibliographystyle{IEEEbib}
\bibliography{icme2023template}

\end{document}